\documentclass[a4paper,twoside]{article}
\usepackage{a4}
\usepackage{amssymb}
\usepackage{amsmath}
\usepackage{amsthm}
\usepackage{mathtools}
\usepackage{authblk}
\usepackage{url}
\usepackage[title]{appendix}
\usepackage{tikz}
\usetikzlibrary{arrows,matrix}

\providecommand{\keywords}[1]
{
	\small	
	\textbf{\textit{Keywords---}} #1
}

\numberwithin{equation}{section}

\title{An Improved Nearest Neighbour Classifier}
\author[1, *]{Eric Setterqvist}
\affil[1]{Johann Radon Institute for Computational and Applied Mathematics (RICAM), Austrian Academy of Sciences, Linz, Austria}
\author[2]{Natan Kruglyak}
\affil[2]{Department of Mathematics, Link\"oping University, Sweden}
\author[3]{Robert Forchheimer}
\affil[3]{Department of Electrical Engineering, Link\"oping University, Sweden, and RISE Research Institutes of Sweden AB}
\affil[*]{Corresponding author: eric.setterqvist@ricam.oeaw.ac.at}

\begin{document}
\maketitle

\begin{abstract}

A windowed version of the Nearest Neighbour (WNN) classifier for images is described. While its construction is inspired by the architecture of Artificial Neural Networks (ANNs), the underlying theoretical framework is based on approximation theory. We illustrate WNN on the datasets MNIST and EMNIST of images of handwritten digits. In order to calibrate the parameters of WNN, we first study it on the classical MNIST dataset. We then apply WNN with these parameters to the challenging EMNIST dataset. It is demonstrated that WNN misclassifies 0.42\% of the images of EMNIST and therefore significantly outperforms predictions by humans and shallow ANNs that both have more than 1.3\% of errors.

\end{abstract}


\section{Introduction} \label{intro}

In spite of their remarkable success in classification problems Artificial Neural Networks (ANNs) work as black boxes and the understanding of their behaviour is limited. Consequently, reasons for their failures are unclear. It is therefore important to continue to develop the traditional classification methods as they typically are easier to interpret and correct when they make mistakes. However, we need to learn from the ANNs how to improve the traditional methods. In particular to incorporate the properties that make ANNs so powerful when it comes to their predictive performance. In this paper we use an approach that is based on the Nearest Neighbour (NN) classifier which belongs to the class of supervised learning algorithms and is based on comparing an unknown object with a set of labelled objects. The label of the closest object (based on some distance metric) is then selected as the classifier result (\cite{Bis06}, pp 124-127).

In our previous paper \cite{KruFor21} we suggested an algorithm which consists of two steps. The first step consists of a windowed version of the NN classifier (Windowed Nearest Neighbour, WNN for short). The main idea behind the WNN algorithm is to compute distances to the object sets over sliding windows of the images resembling convolutional layers of ANNs. In the second step of our algorithm in \cite{KruFor21} we used ANNs as a technical tool to approximately compute the distances to the object sets. The motivation behind these modifications was to make the resulting algorithm firmly rooted in a traditional method but fast enough to be useful in real-life applications. This two step classification algorithm was then applied to two classes of binarised images. However, it was not clear how to extend the algorithm to more general situations and, more important, if it gives good results for difficult classification problems.

In this paper the focus is on the WNN classifier itself without any ANN approximations. Our main goal is to study the WNN classifier on the recently constructed EMNIST dataset of images of handwritten digits \cite{CohAfsTapSch17}. To find parameters of the WNN classifier (size of used windows and parameters of artificial extensions of the training set) we first investigate it in detail on the classical MNIST dataset \cite{mnist}. In particular, as EMNIST is balanced, i.e. the number of images in training sets is the same for different digits, both standard and balanced variants of MNIST are considered.

It is important to note that MNIST and EMNIST both consist of real images which are “normalized” by preprocessing. By “normalized” we mean that the images contain only a single object which is approximately of the same size and in similar position in all images. This is probably the main restriction in the present study of the WNN classifier.

After an overview of related work in Section \ref{sec:rel}, the main features of the WNN classifier are given in Section \ref{wnn}. The performance of the WNN classifier on the MNIST dataset is presented in Section \ref{wnn-results} where we also show the influence of different window sizes and, moreover, the stability of misclassified test images when using different training sets. Section \ref{wnn-extended} describes how the performance can be improved by increasing the number of training images through translations, rotations and certain distortions of the images in the dataset. In Section \ref{emnist}, we present results on the EMNIST dataset. Section \ref{complexity} contains an algorithm for reducing number of used windows and gives complexity estimates of the NN and WNN classifiers. In Section \ref{summary}, a summary and additional side notes are given. Two appendices concludes the paper. Appendix \ref{sec:math} presents the roots of the WNN classifier in approximation theory, while Appendix \ref{sec:tables} contains tables of error rates for different window sizes.

\section{Related Research} \label{sec:rel}
The NN algorithm is one of the most basic classification methods. The algorithm has been known for a long time as the nonparametric alternative to another traditional method - Linear Discriminant Analysis. The NN algorithm as well as a more refined version denoted k-NN, was described and analysed for the first time in 1951 \cite{FixHod51,SilJon89}. Under fairly general assumptions, it was shown in 1966 that NN (also denoted ‘1-NN’ or ‘single NN’) obtains an error rate not larger than twice that of a parametric Bayesian classifier that utilises full knowledge about the underlying statistics \cite{CovHar67}.

There are numerous applications of the NN algorithm to classification of handwritten digits. Of particular interest are those which use the same datasets (MNIST and EMNIST) that we consider in this work. The web page \cite{mnist} gives an overview of results for k-NN and other classifiers applied to MNIST. For the Euclidean norm, an error rate of 3.09\% is reported for k-NN, a result that we have been able to confirm for $\mathrm{k}=1$.

Various methods to artificially extend the training set have been proposed. Most straight-forward is to shift the available training images in order to obtain a larger training set. Using a shift of one pixel in all eight directions, the error rate can be reduced to 2.3\% on MNIST \cite{GroTog19}. Even better results are achieved when local deformations are allowed. Based on independent pixel shifts and an extended distance metric which takes into account local gradients, an error rate of 0.52\% has been reported \cite{KeyDesGolNey07}. To our knowledge, this is the best result published to date for an NN-based classification method on MNIST. It can also be noted that the local context described in that paper bears resemblance with the windowed approach that we have taken in our current work.

The recent survey \cite{BalSaeIsa19} provides a detailed overview of image classification on MNIST and EMNIST, covering both traditional methods and ANNs, where further references can be found.

\section{The WNN Algorithm\label{wnn}}

\subsection{Construction of the WNN algorithm}
The WNN algorithm was introduced in \cite{KruFor21}. It is based on applying the NN algorithm simultaneously on smaller sections (`windows') of the images. This paper investigates the WNN algorithm on the full MNIST dataset and the EMNIST Digits dataset. All images in these datasets are greyscale images of size $28\times 28$, i.e. they consist of 784 pixels and on each pixel the image takes an integer value between $0$ and $255$. For each of the 784 pixels, we will consider a square `window' $W$ centered at the pixel with side length $S$ where $S$ is an odd positive integer. Each window $W$ therefore consists of $S^2$ pixels. For simplicity of notation, we will from now on refer to $S$ as the size of the window. Note that for some pixels, the window $W$ will extend beyond the image boundaries. We will set the values of all pixels in $W$ which falls outside the image equal to zero.

Denote by $A_{train}^{i}$, $i=1,\dots,10$, the class of training images that
corresponds to digit $i-1$. By $B_{test}^{i}$, we denote the corresponding class of test images for digit $i-1$. 
Let $B\in\cup^{10}_{i=1}B_{test}^{i}$ be some test image. We will calculate distances between $B$ and $A_{train}^{i}$ on the window $W$ according to%
\begin{equation}\label{def:dist}
dist_{W}(B,A_{train}^{i})=\min_{A\in A_{train}^{i}}\Big(\sum_{k\in W}(B(k)-A(k))^{2}\Big)^{1/2},\,i=1,\dots,10.
\end{equation}
Next, for each class $A^{i}_{train}$ we take into account the distances on all windows and define%
\begin{equation} \label{def:Dist}
Dist(B,A_{train}^{i})=\Big(\sum_{W}dist_{W}(B,A_{train}^{i})^2\Big)^{1/2},\,i=1,\dots,10.%
\end{equation}
Our algorithm classifies image $B$ as an image from the class
$B_{test}^{i}$ if
\begin{equation*}
Dist(B,A_{train}^{i})=\min_{j\in\left\{1,\dots,10\right\}}Dist(B,A_{train}^{j})\text{.}%
\end{equation*}
In the case when minimum is attained for several indices $1\leq i_1<\dots <i_n\leq 10$, we will (arbitrarily) classify $B$ as an image of $B_{test}^{i_1}$. It can be noted that such a situation has not occurred in our investigations.

Note further that when $S\geq 55$, the WNN classifier coincides with the NN classifier as all windows will cover the image and $dist_W(B,A_{train}^i)$ in this case is the $L^2$-distance between $B$ and $A_{train}^i$.

We next consider two interpretations of the WNN algorithm in terms of, respectively, approximation theory and statistics.

\subsection{Approximation interpretation}
In the first step of the WNN algorithm, we approximate (by using
Euclidean distance) the test image $B$ on the window $W$ by the set of restrictions of images from the class $A_{train}^i$ on $W$ and obtain the \emph{local} distance $dist_W(B,A_{train}^i)$. The second step combines these local distances into a \emph{global} distance $Dist(B,A_{train}^i)$ from $B$ to $A_{train}^i$ and the algorithm predicts, as described above, that $B$ is the digit $i-1$ for which $Dist(B,A_{train}^i)$ is minimal. Some particular notes:
\begin{itemize}
	\item We have used overlapping set of windows and such type of windows appear when characterizing functions with derivatives in the theory of local approximations.
	\item There are different possibilities to combine local distances, we have done experiments and have found that the summation of squared local distances  $dist_W(B,A_{train}^i)^2$ gives the best result.
	\item The global distance $Dist(B,A_{train}^{i})$ resembles a discrete analogue of the description, in terms of local approximations, of the $K$-functional (from real interpolation) for couples of Sobolev spaces.
\end{itemize}
We refer the interested reader to Appendix \ref{sec:math} for further details about the connection of the WNN classifier to approximation theory.

\subsection{Statistical interpretation}
Let a window $W$ with $S^2$ pixels be given and let $X_W$ denote the restriction of an image to $W$. Given a training image $A$, we consider a multivariate normal distribution on $W$ with expected value $A_W$ and independent pixels with uniform standard deviation $\sigma$. For a given test image $B$ we then evaluate the corresponding probability density function
\begin{equation*}
P_A(B_W)=\frac{1}{c}\exp\left(-\frac{1}{2\sigma^2}\sum_{k\in W}(B(k)-A(k))^2\right)
\end{equation*}
where $c=\left(\sqrt{2\pi}\sigma\right)^{S^2}$. In this sense, we interpret restrictions of test images to $W$ as "noisy" restrictions of training images. 

Next, the above situation is carried over to all windows. Labelling the windows by $W_1,W_2,\dots,W_M$, a normally distributed random vector is assigned to $W_j$ having independent pixels with standard deviation $\sigma$ and expected value given by the corresponding restriction of a training image $A_j$. We allow for $A_{j_1}=A_{j_2}$, i.e. the same training image can be used on several windows, and assume further that the random vectors are independent. The resulting joint probability density function can then be evaluated according to
\begin{equation*}
P_{A_1,\dots,A_M}(B_{W_1},\dots,B_{W_M})=\frac{1}{c^M}\exp\left(-\frac{1}{2\sigma^2}\sum_{j=1}^M\sum_{k\in W_j}(B(k)-A_j(k))^2\right).
\end{equation*}

The joint probability density function is now maximized over the different classes $A_{train}^i$, i.e. for $i=1,\dots,10$ we compute
\begin{align*}
P_i(B)&=\max_{A_1,\dots,A_M\in A_{train}^i}P_{A_1,\dots,A_M}(B_{W_1},\dots,B_{W_M})\\
&=\frac{1}{c^M}\exp\left(-\frac{1}{2\sigma^2}\sum_{j=1}^M\min_{A\in A_{train}^i}\sum_{k\in W_j}(B(k)-A(k))^2\right)\\
&=\frac{1}{c^M}\exp\left(-\frac{1}{2\sigma^2}\sum_{j=1}^{M} dist_{W_j}(B,A_{train}^i)^2\right)=\frac{1}{c^M}\exp\left(-\frac{1}{2\sigma^2}Dist(B,A_{train}^i)^2\right).
\end{align*}
Note that $P_{i}(B)$ is maximal when $Dist(B,A_{train}^i)$ is minimal. That is, applying the WNN classifier to $B$ corresponds to choosing the digit $i-1$ which maximizes $P_{i}(B)$. The WNN classifier can therefore be seen as a maximum likelihood estimation of the classes $A_{train}^i$ based on the corresponding training images $A_1,\dots,A_M$.

\section{Experiments on MNIST}
\label{wnn-results}
\begin{figure}
	[ptb]
	\begin{center}
		\includegraphics[
		height=2.7224in,
		width=3.6253in
		]%
		{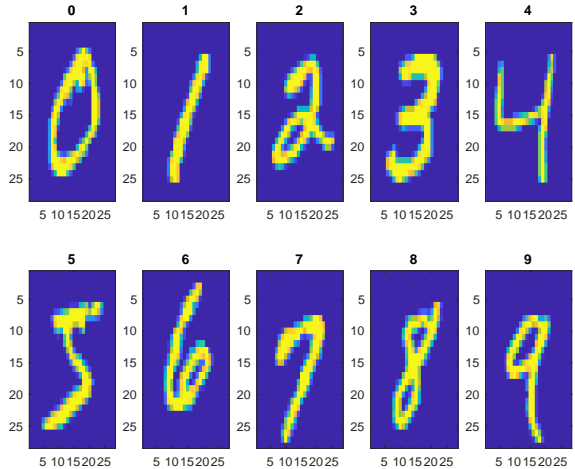}%
	\end{center}
	\caption{10 random MNIST test images. Above each image is its label.}
	\label{fig:10_test_images}
\end{figure}

The MNIST dataset \cite{mnist} contains 60000 greyscale images of handwritten digits $0,1,\dots,9$ which are usually used for training and another 10000 images of handwritten digits which typically are assigned for testing. See Figure \ref{fig:10_test_images} for some examples of test images.

We enumerate the images for each digit in the order they appear in the MNIST dataset, starting with the training set and then the test set. So, for example, for digit $0$ we will have in total $5923$ training images enumerated $1-5923$ and $980$ test images enumerated $5924-6903$, see Table \ref{standard}. Note that there are different number of images in the training and test sets for different digits. In this paper, we will make some modifications of the standard training and test sets. To make the classification of the WNN algorithm more 'balanced' we will use a training set which is uniform in size with respect to different digits. More precisely, the training set will consist of the first $6000$ images for each digit and the test set will consist of all remaining images, see Table \ref{basic}. We will refer to this setup as MNIST Balanced. One needs to keep in mind this difference when comparing performance of classification methods.

\begin{table}[h]
	\centering
	\begin{tabular}{ |c|c|c| } 
		\hline
		Digit & Training & Test\\
		\hline
		0 & 1:5923 & 5924:6903\\
		1 & 1:6742 & 6743:7877\\
		2 & 1:5958 & 5959:6990\\
		3 & 1:6131 & 6132:7141\\
		4 & 1:5842 & 5843:6824\\
		5 & 1:5421 & 5422:6313\\
		6 & 1:5918 & 5919:6876\\
		7 & 1:6265 & 6266:7293\\
		8 & 1:5851 & 5852:6825\\
		9 & 1:5949 & 5950:6958\\
		\hline
		Total & 60000 & 10000\\
		\hline
	\end{tabular}%
	\caption{\label{standard}Standard training and test images, referred to as MNIST Standard.}
\end{table}	

\begin{table}[h]
	\centering
	\begin{tabular}{ |c|c|c| } 
		\hline
		Digit & Training & Test\\
		\hline
		0 & 1:6000 & 6001:6903\\
		1 & 1:6000 & 6001:7877\\
		2 & 1:6000 & 6001:6990\\
		3 & 1:6000 & 6001:7141\\
		4 & 1:6000 & 6001:6824\\
		5 & 1:6000 & 6001:6313\\
		6 & 1:6000 & 6001:6876\\
		7 & 1:6000 & 6001:7293\\
		8 & 1:6000 & 6001:6825\\
		9 & 1:6000 & 6001:6958\\
		\hline
		Total& 60000 & 10000 \\
		\hline
	\end{tabular}%
	\caption{\label{basic}Modified training and test images, referred to as MNIST Balanced.}
\end{table}	

The next table ({Table~\ref{err-basic}}) shows how many classification errors WNN produces for different window sizes $S$ (indicated by WNN$S$, for example by WNN11 we will mean the case when $S=11$ ). By 'NN' we denote the case when $S=55$, i.e. we actually have only one window and our classification algorithm coincides with the usual Nearest Neighbour algorithm. We see that WNN with $S=11$ is the best and has 106 errors from 10000 test images, i.e. the error rate is 1.06\%. From this table we also see that the NN algorithm has 266 errors, i.e. a much higher error rate than the WNN11 algorithm. Note further that the best size of the window ($S=11$) is possible to estimate in advance, see the next subsection.

\begin{table}[h]
	\centering
	\resizebox{1.2\textwidth}{!}{%
		\begin{tabular}{ |c|c|c|c|c|c|c|c|c|c|c|c|c| } 
			\hline
			Digit & NN & WNN3 & WNN5 & WNN7 &WNN9 & WNN11 & WNN13 & WNN15 & WNN17 & WNN19 & WNN21 & WNN23\\ 
			\hline
			0 & 7 & 17 & 7 & 4 & 4 & 5 & 5 & 5 & 5 & 5 & 5 & 5\\ 
			1 & 10 & 123 & 35 & 14 & 6 & 5 & 5 & 7 & 6 & 7 & 7 & 7\\
			2 & 37 & 28 & 13 & 8 & 8 & 7 & 9 & 10 & 12 & 14 & 14 & 16\\
			3 & 48 & 42 & 16 & 16 & 16 & 14 & 12 & 13 & 18 & 19 & 19 & 21\\
			4 & 28 & 14 & 7 & 6 & 5 & 6 & 8 & 8 & 8 & 9 & 12 & 14\\
			5 & 2 & 5 & 1 & 1 & 2 & 2 & 2 & 2 & 2 & 2 & 2 & 2\\
			6 & 12 & 28 & 21 & 11 & 10 & 8 & 7 & 7 & 7 & 7 & 8 & 8\\
			7 & 43 & 76 & 38 & 28 & 19 & 18 & 16 & 20 & 22 & 25 & 26 & 27\\
			8 & 41 & 20 & 9 & 7 & 11 & 12 & 13 & 15 & 13 & 13 & 15 & 15\\
			9 & 38 & 54 & 38 & 31 & 29 & 29 & 30 & 27 & 28 & 29 & 31 & 33\\
			\hline
			Total & 266 & 407 & 185 & 126 & 110 & 106 & 107 & 114 & 121 & 130 & 139 & 148\\
			\hline
		\end{tabular}%
	}
	\caption{\label{err-basic}Errors for different window sizes on MNIST Balanced.}
\end{table}

\subsection{Classification Stability}

\subsubsection{Stability of Window Size}
The window size $S$ is the only parameter of the WNN algorithm. As a demonstration of the stability of the WNN algorithm, we consider different training and test sets on the MNIST dataset and compute the number of errors for different window sizes. If we modify the example of Table \ref{err-basic} and use for each digit in MNIST Balanced the first 5000 images for training and the same test set as before, the window size $S=11$ is again the best, see Table \ref{err_1-5000_6001-YYYY}. Considering further examples, see Tables \ref{err_1-5000_5001-6000} -- \ref{err_1-4000_6001-YYYY} in Appendix \ref{sec:tables}, $S=11$ is the best or second best window size. With these observations at hand, all experiments from now on will use $S=11$.

\begin{table}[h]
	\centering
	\resizebox{1.2\textwidth}{!}{%
		\begin{tabular}{ |c|c|c|c|c|c|c|c|c|c|c|c|c| } 
			\hline
			Digit & NN & WNN3 & WNN5 & WNN7 & WNN9 & WNN11 & WNN13 & WNN15 & WNN17 & WNN19 & WNN21 & WNN23\\ 
			\hline
			0 & 5 & 17 & 5 & 4 & 3 & 3 & 3 & 4 & 4 & 4 & 4 & 4\\ 
			1 & 10 & 127 & 40 & 13 & 9 & 7 & 6 & 8 & 9 & 9 & 8 & 8\\
			2 & 37 & 26 & 14 & 8 & 8 & 9 & 9 & 9 & 11 & 13 & 14 & 15\\
			3 & 47 & 36 & 18 & 18 & 15 & 12 & 14 & 14 & 15 & 18 & 18 & 18\\
			4 & 30 & 16 & 7 & 7 & 6 & 7 & 7 & 7 & 7 & 7 & 9 & 11\\
			5 & 5 & 4 & 2 & 2 & 2 & 2 & 2 & 2 & 2 & 2 & 2 & 2\\
			6 & 13 &  29 & 20 & 13 & 10 & 8 & 7 & 8 & 8 & 7 & 7 & 8\\
			7 & 44 & 77 & 38 & 26 & 20 & 18 & 19 & 22 & 22 & 26 & 25 & 27\\
			8 & 52 & 22 & 7 & 8 & 9 & 15 & 15 & 17 & 17 & 19 & 24 & 24\\
			9 & 42 & 49 & 41 & 31 & 32 & 31 & 30 & 27 & 28 & 31 & 34 & 34\\
			\hline
			Total & 285 & 403 & 192 & 130 & 114 & 112 & 112 & 118 & 123 & 136 & 145 & 151\\
			\hline
		\end{tabular}%
	}
	\caption{\label{err_1-5000_6001-YYYY}Errors for different window sizes. For each digit, training images 1:5000 and test images according to Table \ref{basic} are used.}
\end{table}

\subsubsection{Stability of Misclassified Images} \label{sec:stability}
As we see from Tables \ref{err-basic} and \ref{err_1-5000_6001-YYYY}, WNN11 has on MNIST Balanced 106 errors when images 1:6000 are used for training and 112 errors when images 1:5000 are used for training. Comparing the misclassified images of these two examples shows that 98 of them are common. Defining the fraction of common points for two finite subsets $ A$ and $ B $ by the formula $|A\cap B| /|A\cup B|$, the percentage of common errors in this case is 82{\%} ($|A\cap B|=98$ and $|A\cup B|=120$).

\subsection{Parameter $p$ in the Approximation Norm\label{norm}}

In local approximation, one usually considers approximation in $L^{p}$-norm which leads us to the following generalization of \eqref{def:dist},%

\[
dist_{W,p}(B,A_{i})=\min_{A\in A_{i}}\Big(\sum_{y\in W}\left\vert
B(y)-A(y)\right\vert ^{p}\Big)^{1/p},\,i=1,\dots,10.
\]
We then define
\[
Dist_{p}(B,A_{i})=\Big(\sum_{W}dist_{W,p}(B,A_{i})^{p}\Big)^{1/p}
\]
and classify image $B$ as image of class $i$ if
\[
Dist_{p}(B,A_{i})=\min_{j\in\left\{1,\dots,10\right\}}Dist_{p}(B,A_{j})\text{.}
\]
Experiments with $p=1,3$ were done, see Table \ref{err-L1} and Table \ref{err-L3} in Appendix \ref{sec:tables}. Using the same test and training sets, Table \ref{err_1-5000_5001-6000} in Appendix \ref{sec:tables} shows the outcome of the experiments for $p=2$. Comparing the results gives support to that $p=2$ is a natural choice in terms of predictive performance.

For binarised images the choice of $p$, $1\leq p<\infty$, does not influence the result because we have $\left\vert B(y)-A(y))\right\vert ^{p}=\left\vert B(y)-A(y)\right\vert^{2}$ as $\left\vert B(y)-A(y)\right\vert $ only equals $0$ or $1$. Note that this equality does not hold in general for non-binarised images. We did experiments with binarised images, see {Table~\ref{err-binary}} in Appendix \ref{sec:tables}, and found, as expected, that the result is worse than for non-binarised images (compare with {Table~\ref{err_1-5000_5001-6000}}).

\section{Extending the Training Set of MNIST}\label{wnn-extended}

It is expected that a larger training set will improve the performance of WNN. For this reason, we first extended the training set of MNIST Balanced (denoted `Set 0') artificially by spatially shifting each image not more than one pixel in horizontal, vertical or in both directions at the same time. This gives eight new images for each original training image and in total $60000\cdot 9=540000$ training images. We refer to this set as `Set 1'. In a second step, each training image of Set 1 was rotated $\pm 5,\pm 25$ degrees generating an additional $2.16$ million training images. This set of $2.7$ million training images is denoted `Set 2'. Next, note that the digits are contained in the center $20\times 20$ pixels \cite{mnist} of the image. For each image of Set 1 we then generated four new images by compressing/expanding the width or the height of this center part to $18$ and $22$ pixels. These compressed and expanded images together with Set 1 gives `Set 3' with in total $2.7$ million images. Finally, the images of Sets 2 and 3 together constitute `Set 4' which accordingly contains $4.86$ million unique images. In Table \ref{err_trainingsets}, we give the resulting number of errors of WNN11 for the different training sets.
\begin{table}[h]
	\centering
	\begin{tabular}{ |c|c|c|c|c|c| } 
		\hline
		Training set & Set 0 & Set 1 & Set 2 & Set 3 & Set 4\\ 
		\hline
		No. of training images & 60000 & 540000 & 2700000 & 2700000 & 4860000\\
		\hline
		No. of test images & 10000 & 10000 & 10000 & 10000 & 10000\\
		\hline
		No. of errors on test images & 106 & 62 & 49 & 49 & 41\\ 
		\hline
		Error rate & 1.06\% & 0.62\% & 0.49\% & 0.49\% & 0.41\% \\
		\hline
	\end{tabular}%
	\caption{\label{err_trainingsets}Errors for WNN11 on MNIST Balanced when using different extensions of the training set.}
	
\end{table}
We note that Set 4 gives the lowest error rate with 0.41\%. Further expansions of the training set might give better results but we chose in this study to restrict ourselves to a few straightforward alternatives. When comparing with previously published work on NN-based methods, recall that the training and test sets considered in this study are not the standard ones of MNIST (compare Tables \ref{standard} and \ref{basic}). However, when applying the WNN algorithm (with extension of the training set as above) on MNIST Standard, we obtain an error rate of 0.48\% which is lower than the best published result of 0.52\% \cite{KeyDesGolNey07} that we are aware of. We note that human error rate on MNIST is measured to around 0.2\% \cite{LeCJacBotCorDenDruGuyMueSacSimVap95}.

See Table for a compilation of different results.

\section{EMINST}
\label{emnist}

The EMNIST dataset consists of handwritten digits and letters and was constructed rather recently, see \cite{CohAfsTapSch17}. As for MNIST, it is based on the NIST Special Database 19 and all its images are greyscale and of size $28\times 28$ pixels. The NIST dataset is known to be challenging for classification. For instance, the human error rate has been reported to be more than 1.5\% (see \cite[Fig. 5]{SimLeCDen92}) on this data set. Note that when generating the training and test sets of NIST, two very different populations (census bureau workers generated the training set and high-school students generated the test set) were used, see \cite[p.~57]{SimLeCDen92}. We will focus on the EMNIST Digits subset containing images of handwritten digits with 4000 test images and 24000 training images for each digit $0,1,...,9$. When writing EMNIST below, we refer to this subset.

\begin{figure}[h]
	\begin{center}
		\includegraphics[height=2.7224in, width=3.6253in]{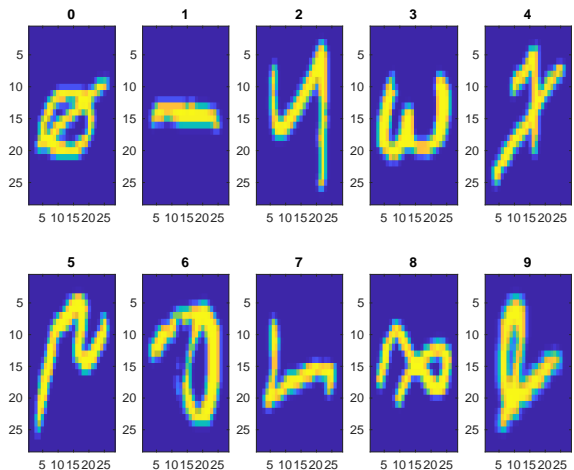}
	\end{center}
	\caption{10 random EMNIST test images. Above each image is its label.}
	\label{fig:10_test_images_EMNIST}
\end{figure}

For the construction of EMNIST, the NIST dataset was preprocessed. In particular, the resulting training and test sets contain samples of both students and census bureau workers, \cite[Sect.~2.B.]{CohAfsTapSch17}. Contrary to MNIST, the images in EMNIST look more like various types of shapes than images of handwritten digits, see e.g. Figure \ref{fig:10_test_images_EMNIST}.

The human error rate on EMNIST might therefore be different than on NIST. We wrote a special program and used it to check human classification errors on EMNIST. It appears that even after a few days of training ourselves, our error rate is more than 1.5\%. We also tried shallow neural networks (with a few hidden layers) and noticed that the error rate was more than 1.3\% for these networks. 
Applying the WNN algorithm on EMNIST, window size 11 will be used based upon previous investigations on MNIST. The original training set of 240000 images is denoted by `Set 0' and the extension of this set by spatially shifting not more than one pixel in horizontal, vertical or in both directions at the same time is referred to as `Set 1'. Adding rotations of $\pm 5,\pm 25$ degrees of each image to this set, a set denoted `Set 2' is constructed. Finally, we consider an extension of Set 0 in terms of a spatial shift of maximum two pixels in horizontal, vertical or in both directions at the same time (denoted `Set 3') and, as in the previous case, then extend Set 3 to include rotated images by $\pm 5,\pm 25$ degrees (denoted `Set 4'). 
These extensions are important. Indeed, the NN algorithm on EMNIST using the training images of Set 0 gives 625 errors (an error rate of 1.56\%)  and 385 errors (an error rate of 0.96\%) using Set 4. The classification results of WNN using these training sets are given in Table \ref{emnist_trainingsets}.

\begin{table}[h]
	\centering
	\begin{tabular}{ |c|c|c|c|c|c| } 
		\hline
		Training set & Set 0 & Set 1 & Set 2 & Set 3 & Set 4\\ 
		\hline
		No. of training images & 240000 & 2160000 & 10800000 & 6000000 & 30000000\\
		\hline
		No. of test images & 40000 & 40000 & 40000 & 40000 & 40000\\
		\hline
		No. of errors on test images & 303 & 195 & 190 & 177 & 168\\
		\hline
		Error rate &  0.76\% & 0.49\% & 0.48\% & 0.44\% & 0.42\%\\
		\hline
	\end{tabular}%
	\caption{\label{emnist_trainingsets}Errors for WNN11 on EMNIST when using different extensions of the training set.}
\end{table}
For comparison, we did experiments with one third of the training images (8000 images per digit) and did extension as for Set 4. The resulting number of errors for the WNN algorithm became 202 giving an error rate of 0.5\%.

The best published result of a traditional classification method applied to EMNIST that we are aware of reports an error rate of 2.26\%, see Table 3 in the survey paper \cite{GhaIngSon18}. Therefore, the result of WNN seems to be state of the art and much better than the human error rate of 1.5\%. Further comparisons with \cite[Table 3]{GhaIngSon18} reveal that the WNN classifier outperforms several ANNs when taking into account extensions of the training set. 

We would like to finish with the following remark. By using a more sophisticated version of the WNN algorithm, as described below, we can reduce the number of errors from 168 to 129 (an error rate of 0.32\%) that is on par with the best neural network result (see \cite{JayJayJayRajSenRod19}). Let us briefly describe this algorithm. Let $A$ be a training image from Set 0. Denote by $A_{ext}$ the set which consists of $A$ and all its extensions as described above (so $A_{ext}$ consists of 125 images). Then we define the distance $d$ from the test image $B$ to $A$ as
\begin{equation*} 
d(B,A)=\Big(\sum_{W}(d_{W}(B,A_{ext}))^2\Big)^{1/2},
\end{equation*}
where $d_W(B,A_{ext})$ is the distance on window W from $B$ to $A_{ext}$ given by
\begin{equation*}
d_W(B,A_{ext})=\min_{X\in A_{ext}}\Big(\sum_{w\in W}(B(w)-X(w))^{2}\Big)^{1/2}.
\end{equation*}
Next, the distance $D$ from $B$ to the training class $A^{i}_{train}$ from Set 0 (containing 24000 images for each digit $i-1$), $i=1,\dots,10$, is defined as
\begin{equation*}  
D(B,A_{train}^{i})=\min_{A\in A^{i}_{train}}d(B,A).
\end{equation*}
We classify $B$ as an image from the class $B_{test}^{i}$ if
\begin{equation*}
D(B,A_{train}^{i})=\min_{j\in\left\{1,\dots,10\right\}}D(B,A_{train}^{j}).
\end{equation*}
This distance algorithm (denoted DWNN) has 148 errors on EMNIST. Now for each test image $B$ we do predictions by WNN and DWNN. Usually these two predictions coincide, if they are different then we classifiy $B$ according to the NN classifier using the two training classes from Set 4 (corresponding to the predictions by WNN and DWNN respectively). The resulting algorithm has 129 errors on EMNIST.

\section{Decreasing Number of Windows and Complexity Estimation} \label{complexity}

\subsection{Algorithm for decreasing number of used windows}

Denote by WNN$_{W_1,...,W_K}$ the WNN algorithm where the windows
$W_{1},...,W_{K}$ are excluded, i.e. instead of $Dist(B,A_{train}^i)$ we use	
\[
Dist_{W_1,\dots,W_K}(B,A_{train}^{i})=\left(\sum dist_{W}(B,A_{train}^{i})^{2}\right)
^{1/2},\,i=1,\dots,10
\]
where we sum over all windows $W$ except $W_{1},...,W_{K}$.

We determine $W_{1},...,W_{K}$ in an iterative way according to the following. First, calculate for each window $W$ the number of errors for the WNN$_{W}$ algorithm. This number is denoted NE$_{W}$ (number of
errors when window $W$ is excluded). We then consider the set of windows $W$ with smallest number of errors NE$_{W}$. Usually this set contains many windows. For every window in the set, consider
\begin{equation*}
GAP_W=\sum_{i=1}^{10}\sum_{B\in B_{test}^{i}}(Dist_{W}(B,A_{train}^{i})-\min_{j\in\left\{1,...,10\right\}}Dist_{W}(B,A_{train}^{j}))
\end{equation*}
and exclude the window $W$ for which GAP$_{W}$ is maximal. To explain the idea behind this algorithm note that if the test image $B$
is in class $B_{test}^i$ and is predicted correctly by WNN then
\[
Dist_{W}(B,A_{train}^{i})-\min_{j=1,...,10}Dist_{W}(B,A_{train}^{j})=0
\]
and if $B$ is not predicted correctly then
\[
Dist_{W}(B,A_{train}^{i})-\min_{j=1,...,10}Dist_{W}(B,A_{train}^{j})>0.
\]
So if all test images are predicted correctly then $GAP_{W\text{ }}=0$ and
idea is to exclude the worst window.

\begin{figure}[h]
	\begin{center}
		\includegraphics[
		height=2.7224in,
		width=3.6253in
		]%
		{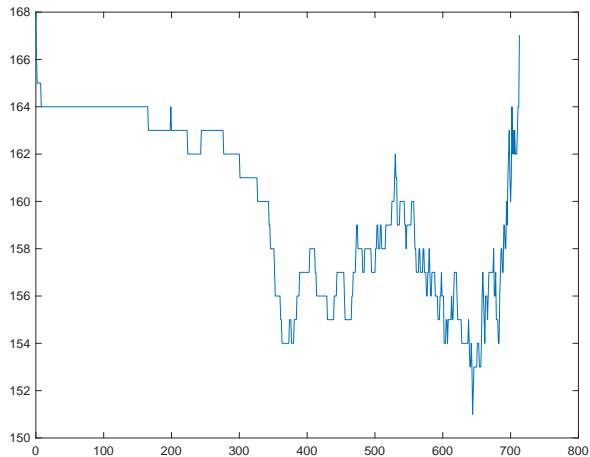}%
	\end{center}
	\caption{Number of errors versus number of excluded windows.}
	\label{fig:error_excl_win_1}
\end{figure}

Let us next consider an example on EMNIST. We apply the above algorithm to Set 4 (recall that it consists of 30$\times10^{6}$ images). Then the number of errors with respect to the number of excluded windows $K$ can be seen in Figure \ref{fig:error_excl_win_1}.
In particular when 100 windows are used, i.e. $K=684$, the number of errors
will be 156. Using only 60 windows the number of errors increase to 167 which still is less than 168 which was obtained using all 784 windows. 

\begin{figure}[h]
	\begin{center}
		\includegraphics[
		height=2.7224in,
		width=3.6253in
		]%
		{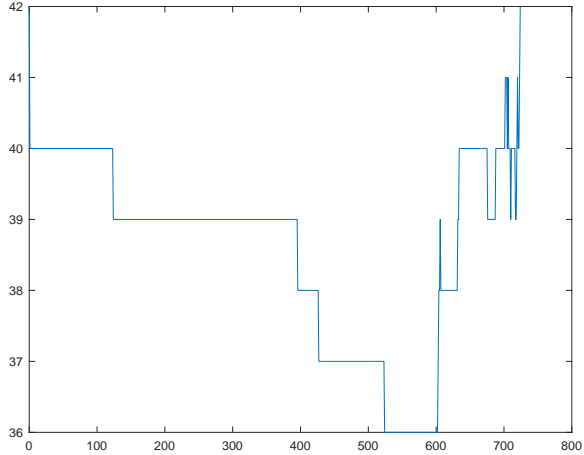}%
	\end{center}
	\caption{Number of errors versus number of excluded windows using validation set.}
	\label{fig:error_excl_win_2}
\end{figure}

However, note that constructing the set of excluded windows by using the whole test set is not correct. We therefore divide the test set of 40000 images randomly into two subsets: the validation set of 30000 images (3000 images for each digit) will be used for determining the number of excluded windows and the remaining set of 10000 images will be the new test set. The resulting graph of errors can be seen in Figure \ref{fig:error_excl_win_2}.%
In particular, if we use just 50 windows then the number of errors will be 42 corresponding to an error rate of 0.42\%. Recall that this error rate is the same as when using all 784 windows on the original test set.

\subsection{Algorithm complexity}
Algorithm complexity can be estimated in different ways. Here, we will count the number of elementary arithmetic and logical operations. This measure indicates the effort in time and/or hardware requirement that is needed to execute the algorithms. To simplify, we regard addition, multiplication, comparison and table look-up to have equal weight.

\subsubsection{NN algorithm}
We need to compute the squared distance between the input image $B$ and each training image. For every pixel, one difference, one squaring and one accumulation operation are required and these operations have to be repeated for every training image. Moreover, we need to find the closest training image, with respect to the squared distance, to $B$. Assuming we have $M$ training images per digit class results in
\begin{equation*}
10\cdot M\cdot(784\cdot 3+1)
\end{equation*}
operations per classification. For $M=24000$, we then have $5.6\cdot 10^8$ operations per classification.

\subsubsection{WNN algorithm}
Here we need to compute the squared distance between windows taken from the input image $B$ and corresponding windows from each training image. For each pixel within such a window this will require one difference operation, one squaring and one accumulation resulting in $3S^2$ elementary operations given a window size $S$. Setting the number of training images per digit to $M$, the next step is to repeat this calculation $M$ times followed by finding the minimum of the squared distances. Thus, for a specific window and digit class, this results in $M\cdot(3S^2 + 1)$ operations. The sum of all $784$ such window distances give the total distance value for a specific digit class. Finally, these calculations are performed for all the $10$ digits and the minimum is registered. This requires altogether

\begin{equation*}
10\cdot M\cdot (784\cdot(3S^2 + 1)+1)
\end{equation*}
operations. Inserting $M = 24000$ and $S = 11$ results in $6.8\cdot 10^{10}$ operations per classification. So, the WNN classifier requires two orders of magnitude more operations per classification than NN.

\section{Summary and Conclusions} \label{summary}
We have used a windowed version of the Nearest Neighbour classifier and applied it to the MNIST dataset of handwritten digits. The performance for different window sizes and different distance metrics is reported. It is shown that there is good stability with regards to the best window size when the training and test sets are changed. This is promising as it opens up the possibility to use part of the training data to search for the best window size. The training set of MNIST was then extended through shifts, rotations and non-uniform scalings. The error rate is decreased substantially when using the extended training sets giving superior performance.

With parameters determined by investigations on MNIST, the WNN classifier was then applied to the EMNIST dataset. Like for MNIST, extensions of the training set greatly improves the classification performance. However, even without such extensions the WNN classifier achieves state of the art results for traditional methods and also performs much better than human prediction.

An algorithm was described for reducing the number of windows. With this algorithm, we could decrease the computational cost of the WNN classifier with one order of magnitude while keeping its predictive performance.

We will end by giving one side note. A series of experiments were done with a k-Nearest Neighbour (k-NN) version of the WNN classifier. However, we did not achieve any improvements in performance. It seems that the local approximation on windows in the WNN classifier already generates a robustness towards outliers in the data.

\newpage

\begin{appendices}

\section{Approximation roots of the WNN algorithm} \label{sec:math}

\subsection{Local approximation and a result in real interpolation} \label{sec:approximation}

We recall that the modern theory of local approximations was developed in the
1970s by Yu. Brudnyi (see \cite{brudnyi}) who used them to describe spaces of differentiable
functions by their local approximations by polynomials of fixed degree. Let us
formulate one result from Brudnyi's theory.

Let $f\in L^{p}(Q_{0}),1<p<\infty$ and $Q_{0}$ be a cube in
$\mathbb{R}^{n},$ here and everywhere below we suppose that cube faces are parallel to the coordinate hyperplanes. Let $Q$ be a cube in $\mathbb{R}^{n}$ with center in $Q_{0}$.
Then the quantity%
\begin{equation}
E_{k}(f,Q)_{p}=\inf_{P}(\int_{Q\cap Q_{0}}\left\vert f(x)-P(x)\right\vert
^{p}dx)^{1/p},\label{1}%
\end{equation}
where infimum is taken over all polynomials $P$ of degree strictly less than
$k$ is called a local approximation of function $f$. Let us consider the well-known
in approximation theory $k$-modulus of continuity%
\[
\omega_{k}(f,t)_{p}=\sup_{\left\vert h\right\vert <t}\left\Vert \sum_{j=0}%
^{k}(-1)^{k-j}\frac{k!}{j!(k-j)!}f(x+jh)\right\Vert _{L^{p}},
\]
where sup is taking over all $h\in\mathbb{R}^{n}$ such that $\left\vert
h\right\vert <t$ and $x,x+h,...,x+kh\in Q_{0}$.

In \cite{brudnyi} Brudnyi showed that with constants of equivalence independent of $f$ and $t>0$
we have%
\[
\omega_{k}(f,t)_{p}\approx\sup_{\left\{  Q_{i}\right\}  }\left(  \sum_{Q_{i}%
}(E_{k}(f,Q_{i})_{p})^{p}\right)  ^{1/p},
\]
where sup is taken over all finite families $\left\{Q_{i}\right\}$ of
cubes $Q_{i}$ with centers in $Q_{0}$, side length equal to $t$ and disjoint
interiors. It was indicated by Peetre \cite{Pee68} that the modulus of continuity $\omega
_{k}(f,t)_{p}$ is deeply connected with the $K$-functional of real
interpolation. Brudnyi showed later in \cite{brudnyi} that for the couple $(L^{p},\dot{W}%
_{p}^{k})$ on the cube $Q_{0}$, where $\dot{W}_{p}^{k}$ is a homogenous Sobolev
space defined by finiteness of the quasinorm
\[
\left\Vert f\right\Vert _{\dot{W}_{p}^{k}}=\sup_{k_{1}+...+k_{n}=k}\left\Vert
\frac{\partial^{k_{1}}}{\partial^{k_{1}}x_{1}}...\frac{\partial^{k_{n}}%
}{\partial^{k_{n}}x_{n}}f\right\Vert _{L^{p}},
\]
the $K$-functional%
\[
K(t,f,L^{p},\dot{W}_{p}^{k})=\inf_{g\in\dot{W}_{p}^{k}}(\left\Vert
f-g\right\Vert _{L^{p}}+t\left\Vert g\right\Vert _{\dot{W}_{p}^{k}}),\text{
	\ }t>0,
\]
is equivalent to the modulus of continuity
\[
K(t^{k},f,L^{p},\dot{W}_{p}^{k})\approx\omega_{k}(f,t)_{p}%
\]
with constants of equivalence independent of $f$ and $t$. So, the
$K$-functional of the couple $(L^{p},\dot{W}_{p}^{k})$ can be described in
terms of local approximations
\[
K(t^{k},f,L^{p},\dot{W}_{p}^{k})\approx\sup_{\left\{  Q_{i}\right\}  }\left(
\sum_{Q_{i}}(E_{k}(f,Q_{i})_{p})^{p}\right)  ^{1/p},
\]
where sup is taken over all finite families $\left\{Q_{i}\right\}$ of
cubes $Q_{i}$ centered in $Q_{0}$ with side length equal to $t$ and disjoint interiors.

Later, see \cite[Thm.~9.2]{KisKru13}, expressions (in terms of local approximations) were given 
for $K$-functionals of the couples $(L^{p_{0}},\dot{W}_{p_{1}}^{k})$. In
particular, a different formula for the $K$-functional of the couple
$(L^{p},\dot{W}_{p}^{k})$ for functions on $\mathbb{R}^{n}$ was found. To formulate it
let us split $\mathbb{R}^{n}$ on cubes $Q_{i}$ with side length equal to $t$ and
consider another family of cubes $\left\{\Omega_{i}\right\}$ where $\Omega_{i}$ has
the same center as $Q_{i}$ but side length $\frac{3}{2}t$ (note that
neighbouring cubes $\Omega_{i}$ and $\Omega_{j}$ intersect). Then%
\begin{equation}
K(t^{k},f,L^{p},\dot{W}_{p}^{k})\approx\left(  \sum_{\Omega_{i}}(E_{k}(f,\Omega_{i}%
)_{p})^{p}\right)  ^{1/p}=\left(  \sum_{\Omega_{i}}\inf_{\deg(P)<k}(\int_{\Omega_{i}%
}\left\vert f(x)-P(x)\right\vert ^{p}dx)\right)^{1/p}.\label{2}%
\end{equation}
That is, we do not need to take supremum over all finite families $\left\{
Q_{i}\right\}$ with side length $t$ and disjoint interiors.

The right hand side of (\ref{2}) suggests the classification algorithm considered in the next subsection.

\subsection{A Classification Algorithm based on Local Approximations}

Let $A_{1},...,A_{M}$ be some sets in $L^{p}$. Then for any cube $Q$ with
center in $Q_{0}$  we can define
$M$ local approximations according to%
\[
E_{A_{m}}(f,Q)_{p}=\inf_{g\in A_{m}}(\int_{Q\cap Q_{0}}\left\vert
f(x)-g(x)\right\vert ^{p}dx)^{1/p},\text{ \ }m=1,...,M.
\]
Note that above, in subsection \ref{sec:approximation}, we consider the case when $M=1$ and $A_{1}$ is the set of
polynomials of degree strictly less than $k$. Suppose also that some family
$\left\{  W_{i}\right\}  $ of cubes ("windows") with centers in $Q_{0}$ and equal side length are given. Then the right hand side of (\ref{2}) suggests to
consider $M$ "distances"%
\[
Dist(f,A_{m})=\left(  \sum_{W_{i}}(E_{A_{m}}(f,W_{i})_{p})^{p}\right)
^{1/p},\text{ \ }m=1,...,M.
\]
Our classification algorithm classifies $f$ as from class $m$ if%
\[
Dist(f,A_{m})=\min_{j=1,...,M}Dist(f,A_{j}).
\]
In the case when minimum is attained for several indices $1\leq j_{1}%
<...<j_{n}\leq M$ we will (arbitrarily) classify $f$ as an image of the class
$A_{j_{1}}$.

\subsection
{Connections to Nearest Neighbour classifier and Artificial Neural Networks}

Suppose that we have several classes of labeled images $A_{1},...,A_{M}$ and
$B$ is an image that we need to classify. Note, that grey images can be
considered as a function  defined on the
set of discrete points (pixels) in $\mathbb{R}^{2}$ (for colour images we need to consider three functions). We will suppose that
pixels are all points with integer coordinates and functions that corresponds
to images are equal to zero for pixels outside the screen, i.e. outside some
fixed cube that we denote by $Q_{0}$. Note that for colour images number of pixels on each window will be three times more. 

In the NN algorithm we calculate distances from $B$ to each class $A_{j}$, 
$j=1,...,M$, and classify $B$ as an element from the class $A_{i}$ if the distance from
$B$ to $A_{i}$ is the smallest. So, the algorithm in subsection 2.2
coincides with the NN classification in the case when the set of cubes (windows)
$\left\{  W_{i}\right\}  $ consists of just one window $W=Q_{0}$. Our classification algorithm based on local approximations can therefore be considered as a windowed version of the NN classifier.

Connections with ANN classifiers are more complicated. We first note that standard feedforward neural networks with ReLU activation
function can be considered as consecutively applying the following two
parts. The first part $F_{1}$ is nonlinear and transform image $B$ that we
need to classify to some space $\mathbb{R}^{N}$ while the second part $F_{2}$ is
linear and transform $\mathbb{R}^{N}$ to $\mathbb{R}^{M}$ where $M$ is the number
of classes. Then $B$ is classified as an element of class $A_{j}$ if
coordinate $j$ is maximal in $F_{2}F_{1}(B)$. Moreover, it is possible to
prove that the nonlinear part $F_{1}$ can be seen as several (sometimes more
than hundred) consecutively applied convolution transformations $T_{k}$ with
ReLU activation function%
\[
F_{1}(B)=T_{K}(T_{K-1}(\dots(T_{1}(B))\dots))
\]
where by convolution transformation with ReLU activation function we mean the
following transformation. Let $\left\{  W_{i}\right\}
_{i=1,...,I}$ be some set of windows with equal size, i.e. each window contains the same number of pixels which we denote by $n$,
and let $L:\mathbb{R}^{n}\rightarrow
\mathbb{R}^{N}$ be some linear map with $L(x)_{+}=\max(L(x),0)$. Then by
convolution transform with ReLU activation function we mean the nonlinear
transform
\[
T(B)=(L_{+}(B_{W_{1}}),...,L_{+}(B_{W_{I}})),
\]
where $B_{W}$ is the restriction of image $B$ to the pixels from window $W$. The name
"convolution" corresponds to the property that on all windows the transformations
$L_{+}$ are the same.

Comparing with our classification algorithm based on local approximations, instead of the nonlinear mapping $F_{1}$ we consider on each window $W_{i}$ the quantity 
\begin{equation*}
((E_{A_{1}}(f,W_{i})_{p})^{p},\dots,(E_{A_{M}}(f,W_{i})_{p})^{p})
\end{equation*}
which plays the role of $L_{+}(B_{W_{i}})$. Next, the counterpart of the linear mapping $F_{2}$ is the summation over all windows for each class $A_{m}$ according to
\[
Dist(f,A_{m})^{p}=\sum_{W_{i}}(E_{A_{m}}(f,W_{i})_{p})^{p}.%
\]
Note further that, similarly to $L_{+}$, the formulas for local approximations $E_{A_{m}}(f,\cdot)_{p}$ are the same for different windows and also use restrictions of $f$ on used windows.

\section{Tables} \label{sec:tables}
This section collects tables referred to in the main part of the paper. For detailed descriptions, see the captions of the individual tables.

\begin{table}[h]
	\centering
	\resizebox{1.2\textwidth}{!}{%
		\begin{tabular}{ |c|c|c|c|c|c|c|c|c|c|c|c|c| } 
			\hline
			Digit & NN & WNN3 & WNN5 & WNN7 & WNN9 & WNN11 & WNN13 & WNN15 & WNN17 & WNN19 & WNN21 & WNN23\\ 
			\hline
			0 & 8 & 25 & 8 & 5 & 6 & 6 & 5 & 5 & 4 & 5 & 5 & 5\\ 
			1 & 6 & 92 & 37 & 18 & 10 & 7 & 6 & 5 & 5 & 5 & 4 & 3\\
			2 & 27 & 15 & 4 & 4 & 6 & 6 & 7 & 8 & 8 & 8 & 11 & 14\\
			3 & 36 & 42 & 16 & 11 & 10 & 11 & 12 & 13 & 15 & 16 & 16 & 17\\
			4 & 37 & 31 & 16 & 12 & 14 & 16 & 17 & 19 & 20 & 20 & 20 & 21\\
			5 & 49 & 53 & 24 & 15 & 17 & 18 & 16 & 19 & 21 & 21 & 21 & 22\\
			6 & 4 & 27 & 14 & 8 & 3 & 2 & 2 & 2 & 2 & 3 & 3 & 4\\
			7 & 24 & 58 & 21 & 8 & 8 & 10 & 11 & 11 & 9 & 10 & 11 & 12\\
			8 & 56 & 30 & 20 & 16 & 16 & 19 & 24 & 23 & 25 & 24 & 24 & 29\\
			9 & 46 & 58 & 42 & 32 & 28 & 31 & 34 & 36 & 36 & 37 & 37 & 36\\
			\hline
			Total & 293 & 431 & 202 & 129 & 118 & 126 & 134 & 141 & 145 & 149 & 152 & 163\\
			\hline
		\end{tabular}%
	}
	\caption{\label{err_1-5000_5001-6000}Errors for different window sizes. For each digit, training images 1:5000 and test images 5001:6000 are used.}
\end{table}

\begin{table}[h]
	\centering
	\resizebox{1.2\textwidth}{!}{%
		\begin{tabular}{ |c|c|c|c|c|c|c|c|c|c|c|c|c| } 
			\hline
			Digit & NN & WNN3 & WNN5 & WNN7 & WNN9 & WNN11 & WNN13 & WNN15 & WNN17 & WNN19 & WNN21 & WNN23\\ 
			\hline
			0 & 10 & 28 & 10 & 6 & 6 & 6 & 5 & 4 & 4 & 4 & 4 & 4\\ 
			1 & 7 & 75 & 27 & 16 & 13 & 10 & 8 & 8 & 7 & 6 & 5 & 5\\
			2 & 47 & 21 & 13 & 13 & 14 & 15 & 16 & 19 & 24 & 27 & 29 & 32\\
			3 & 57 & 60 & 38 & 35 & 33 & 34 & 32 & 30 & 33 & 35 & 36 & 36\\
			4 & 44 & 30 & 9 & 7 & 8 & 6 & 7 & 10 & 13 & 17 & 17 & 21\\
			5 & 39 & 34 & 14 & 7 & 6 & 6 & 7 & 9 & 10 & 15 & 17 & 20\\
			6 & 14 & 29 & 12 & 8 & 8 & 8 & 8 & 8 & 8 & 9 & 9 & 10\\
			7 & 37 & 72 & 28 & 22 & 20 & 17 & 20 & 21 & 23 & 22 & 23 & 24\\
			8 & 77 & 29 & 22 & 20 & 22 & 29 & 31 & 33 & 39 & 41 & 45 & 50\\
			9 & 42 & 63 & 41 & 30 & 25 & 27 & 24 & 19 & 21 & 21 & 25 & 29\\
			\hline
			Total & 374 & 441 & 214 & 164 & 155 & 158 & 158 & 161 & 182 & 197 & 210 & 231\\
			\hline
		\end{tabular}%
	}
	\caption{\label{err_1-4000_4001-5000}Errors for different window sizes. For each digit, training images 1:4000 and test images 4001:5000 are used.}
	
\end{table}

\begin{table}[h]
	\centering
	\resizebox{1.2\textwidth}{!}{%
		\begin{tabular}{ |c|c|c|c|c|c|c|c|c|c|c|c|c| } 
			\hline
			Digit & NN & WNN3 & WNN5 & WNN7 & WNN9 & WNN11 & WNN13 & WNN15 & WNN17 & WNN19 & WNN21 & WNN23\\ 
			\hline
			0 & 9 & 17 & 7 & 5 & 4 & 4 & 5 & 5 & 4 & 4 & 4 & 6\\ 
			1 & 6 & 67 & 37 & 19 & 13 & 8 & 8 & 7 & 5 & 5 & 5 & 3\\
			2 & 35 & 17 & 10 & 7 & 7 & 6 & 7 & 7 & 10 & 10 & 12 & 15\\
			3 & 39 & 37 & 20 & 12 & 12 & 12 & 14 & 14 & 14 & 15 & 16 & 19\\
			4 & 38 & 26 & 15 & 11 & 12 & 13 & 14 & 16 & 17 & 17 & 17 & 19\\
			5 & 55 & 49 & 22 & 18 & 19 & 22 & 22 & 22 & 22 & 23 & 26 & 29\\
			6 & 4 & 23 & 13 & 9 & 9 & 6 & 4 & 3 & 3 & 3 & 3 & 4\\
			7 & 28 & 54 & 24 & 16 & 12 & 12 & 11 & 12 & 12 & 13 & 14 & 12\\
			8 & 64 & 36 & 24 & 22 & 17 & 17 & 23 & 24 & 23 & 27 & 31 & 37\\
			9 & 48 & 62 & 50 & 40 & 34 & 33 & 31 & 35 & 37 & 38 & 39 & 39\\
			\hline
			Total & 326 & 388 & 222 & 159 & 139 & 133 & 139 & 145 & 147 & 155 & 167 & 183\\
			\hline
		\end{tabular}%
	}
	\caption{\label{err_1-4000_5001-6000}Errors for different window sizes. For each digit, training images 1:4000 and test images 5001:6000 are used.}
	
\end{table}

\begin{table}[h]
	\centering
	\resizebox{1.2\textwidth}{!}{%
		\begin{tabular}{ |c|c|c|c|c|c|c|c|c|c|c|c|c| } 
			\hline
			Digit & NN & WNN3 & WNN5 & WNN7 & WNN9 & WNN11 & WNN13 & WNN15 & WNN17 & WNN19 & WNN21 & WNN23\\ 
			\hline
			0 & 6 & 19 & 6 & 3 & 3 & 4 & 3 & 4 & 4 & 4 & 4 & 4\\ 
			1 & 10 & 127 & 39 & 15 & 8 & 6 & 6 & 7 & 8 & 9 & 9 & 9\\
			2 & 37 & 31 & 16 & 10 & 9 & 9 & 10 & 12 & 14 & 15 & 17 & 17\\
			3 & 49 & 38 & 21 & 20 & 15 & 14 & 14 & 15 & 18 & 18 & 18 & 19\\
			4 & 30 & 17 & 9 & 9 & 7 & 8 & 8 & 8 & 7 & 9 & 10 & 10\\
			5 & 6 & 2 & 2 & 2 & 2 & 2 & 2 & 3 & 3 & 3 & 2 & 2\\
			6 & 14 & 31 & 20 & 11 & 10 & 8 & 9 & 9 & 10 & 10 & 10 & 10\\
			7 & 43 & 81 & 43 & 24 & 20 & 20 & 25 & 26 & 29 & 29 & 28 & 29\\
			8 & 58 & 16 & 10 & 9 & 11 & 14 & 15 & 20 & 19 & 23 & 24 & 27\\
			9 & 41 & 57 & 39 & 31 & 35 & 30 & 30 & 30 & 31 & 35 & 36 & 36\\
			\hline
			Total & 294 & 419 & 205 & 134 & 120 & 115 & 122 & 134 & 143 & 155 & 158 & 163\\
			\hline
		\end{tabular}%
	}
	\caption{\label{err_1-4000_6001-YYYY}Errors for different window sizes. For each digit, training images 1:4000 and test images according to Table \ref{basic} are used.}
	
\end{table}

\begin{table}[h]
	
	\begin{tabular}{ |c|c|c|c|c|c|c|c| } 
		\hline
		Digit & NN & WNN5 & WNN7 & WNN9 & WNN11 & WNN13 & WNN15\\ 
		\hline
		0 & 9 & 7 & 6 & 6 & 5 & 5 & 5\\ 
		1 & 6 & 31 & 15 & 11 & 7 & 6 & 4\\
		2 & 32 & 6 & 6 & 6 & 6 & 7 & 9\\
		3 & 40 & 18 & 13 & 13 & 14 & 15 & 14\\
		4 & 44 & 17 & 13 & 16 & 17 & 19 & 19\\
		5 & 55 & 24 & 17 & 17 & 20 & 20 & 22\\
		6 & 4 & 14 & 8 & 4 & 2 & 2 & 2\\
		7 & 27 & 24 & 14 & 12 & 10 & 9 & 8\\
		8 & 75 & 25 & 21 & 19 & 25 & 27 & 32\\
		9 & 61 & 48 & 35 & 33 & 37 & 38 & 40\\
		\hline
		Total & 353 & 214 & 148 & 137 & 143 & 148 & 155\\
		\hline
	\end{tabular}
	\caption{\label{err-L1}Errors for different window sizes and $L^1$-distance. For each digit, training images 1:5000 and test images 5001:6000 are used.}
\end{table}

\begin{table}[h]
	\begin{tabular}{ |c|c|c|c|c|c|c|c| } 
		\hline
		Digit & NN & WNN5 & WNN7 & WNN9 & WNN11 & WNN13 & WNN15\\ 
		\hline
		0 & 6 & 7 & 6 & 4 & 5 & 5 & 5\\ 
		1 & 6 & 38 & 18 & 12 & 11 & 8 & 7\\
		2 & 23 & 5 & 5 & 8 & 7 & 7 & 7\\
		3 & 37 & 22 & 11 & 10 & 11 & 10 & 11\\
		4 & 35 & 15 & 10 & 11 & 12 & 14 & 15\\
		5 & 47 & 21 & 17 & 18 & 19 & 18 & 19\\
		6 & 4 & 13 & 9 & 9 & 5 & 4 & 3\\
		7 & 19 & 21 & 14 & 10 & 10 & 10 & 11\\
		8 & 48 & 21 & 20 & 18 & 19 & 20 & 22\\
		9 & 44 & 49 & 43 & 37 & 36 & 33 & 33\\
		\hline
		Total & 269 & 212 & 153 & 137 & 135 & 129 & 133\\
		\hline
	\end{tabular}
	\caption{\label{err-L3}Errors for different window sizes and $L^3$-distance. For each digit, training images 1:5000 and test images 5001:6000 are used.}
\end{table}

\begin{table}[h]
	\begin{tabular}{ |c|c|c|c|c|c|c|c| } 
		\hline
		Digit & NN & WNN5 & WNN7 & WNN9 & WNN11 & WNN13 & WNN15\\ 
		\hline
		0 & 11 & 13 & 6 & 5 & 5 & 6 & 6\\ 
		1 & 7 & 53 & 25 & 11 & 9 & 8 & 6\\
		2 & 55 & 14 & 8 & 9 & 8 & 13 & 14\\
		3 & 48 & 26 & 21 & 20 & 19 & 18 & 19\\
		4 & 56 & 24 & 17 & 17 & 19 & 23 & 25\\
		5 & 60 & 29 & 22 & 22 & 23 & 22 & 24\\
		6 & 7 & 20 & 11 & 4 & 2 & 2 & 3\\
		7 & 34 & 35 & 19 & 15 & 14 & 16 & 15\\
		8 & 107 & 21 & 26 & 25 & 27 & 39 & 38\\
		9 & 81 & 51 & 46 & 43 & 43 & 46 & 48\\
		\hline
		Total & 466 & 286 & 201 & 171 & 169 & 193 & 198\\
		\hline
	\end{tabular}
	\caption{\label{err-binary}Binarised training and test images. Errors for different window sizes. For each digit, training images 1:5000 and test images 5001:6000 are used.}
\end{table}

\end{appendices}

\clearpage
\pagebreak

\noindent
\textbf{Acknowledgements} \hspace{0.1cm} We acknowledge computational resources from the National Supercomputer Centre at Linköping University through the project Nearest Neighbour Classifier.

\section*{Declarations}

\noindent
\textbf{Funding} \hspace{0.1cm} This work was supported by computational resources from the National Supercomputer Centre at Linköping University.

\vspace{0.3cm}
\noindent
\textbf{Availability of data and material} \hspace{0.1cm} The data sets used in this work are publicly available as reported in Section \ref{wnn-results} and Section \ref{emnist}.

\vspace{0.3cm}
\noindent
\textbf{Code availability} \hspace{0.1cm} Implementations in MATLAB of the image classifiers described in this work are available upon request.

\vspace{0.3cm}
\noindent
\textbf{Conflicts of interest/Competing interests} \hspace{0.1cm} The authors declare that they have no confict of interest or competing interests.

\vspace{0.3cm}
\noindent
\textbf{Ethics approval} \hspace{0.1cm} Not applicable.

\vspace{0.3cm}
\noindent
\textbf{Consent to participate} \hspace{0.1cm} Not applicable.

\vspace{0.3cm}
\noindent
\textbf{Consent for publication} \hspace{0.1cm} Not applicable.

\vspace{0.3cm}
\noindent
\textbf{Authors' contributions} \hspace{0.1cm} E.S. and N.K. did computer implementations and experiments, all other aspects of the paper have been done in collaboration between all authors.

\bibliographystyle{apalike} 
\bibliography{ref} 

\end{document}